\documentclass[final,5p,times,twocolumn]{elsarticle}

\usepackage{subcaption}
\usepackage{graphicx}
\usepackage{multirow}
\usepackage{epsfig}
\usepackage{amsmath}

\usepackage{amssymb,soul,comment}
\usepackage{mathtools}
\usepackage{xcolor}
\usepackage{soul}

\usepackage[switch]{lineno} 
\usepackage{url}
\usepackage{dirtytalk}

\journal{Machine Learning with Applications}
\setcitestyle{square}

\newcommand{\SystemName}{DA-FDFtNet}
\begin{document}

\begin{frontmatter}






\title{DA-FDFtNet:  Dual Attention Fake Detection Fine-tuning Network to Detect Various AI-Generated Fake Images}


\author[1]{Young Oh Bang}
\author[2]{Simon S. Woo\corref{mycorrespondingauthor}}

\cortext[mycorrespondingauthor]{Corresponding author}
\ead{swoo@g.skku.edu}

\address[1]{Department of Artificial Intelligence, Sungkyunkwan University, Suwon, South Korea}
\address[2]{Department of Applied Data Science, Computer Science \& Engineering Department, Sungkyunkwan University, Suwon, South Korea}


\begin{abstract}
Due to the advancement of Generative Adversarial Networks (GAN), Autoencoders, and other AI technologies, it has been much easier to create fake images such as \say{Deepfakes.} More recent research has introduced few-shot learning, which uses a small amount of training data to produce fake images and videos more effectively. Therefore, the ease of generating manipulated images and the difficulty of distinguishing those images can cause a serious threat to our society, such as propagating fake information. However, detecting realistic fake images generated by the latest AI technology is challenging due to the reasons mentioned above. In this work, we propose Dual Attention Fake Detection Fine-tuning Network (\SystemName) to detect the manipulated fake face images from the real face data. Our \SystemName~integrates the pre-trained model with \textit{Fine-Tune Transformer}, MBblockV3, and a \textit{channel attention module} to improve the performance and robustness across different types of fake images. In particular, Fine-Tune Transformer consists of multiple numbers of an image-based self-attention module and a down-sampling layer. \textit{The channel attention module} is also connected with the pre-trained model to capture the fake images feature space. We experiment with our~\SystemName~with the FaceForensics++ dataset and various GAN-generated datasets, and we show that our approach outperforms the previous baseline models.
\end{abstract}


\end{frontmatter}



\section{Introduction}
Since the introduction of Generative Adversarial Networks (GANs)\cite{goodfellow2014generative} and various image forgery techniques~\cite{karras2017progressive,zakharov2019few,karras2019style,wu2019sliced}, various methods~\cite{karras2017progressive, karras2019style, wu2019sliced, choi2019stargan, karras2020analyzing, shaham2019singan} can produce highly realistic images with human faces, scenery, and objects.
Moreover, recent approaches~\cite{zakharov2019few, sun2019meta} using few-shot learning, a technique based on GAN, allow deep learning models to produce high-quality outputs with only a small amount of training data. Generation of such images using few-shot learning can be also exploited for creating fake images, where the recent cases~\cite{cnn_business, yin_2019, deepnude} demonstrate the misuse of those images for malicious purposes. For detecting image forgeries, earlier research leveraged the metadata information or handcrafted characteristics of images. However, the latest AI methods, such as GAN, generate new images as well as metadata from scratch. Therefore, previous metadata-based detection approaches are not practical and useful against recent fake images generated from GANs. In order to address these issues, Convolutional Neural Networks (CNNs)-based binary classifiers such as \textit{ShallowNet}~\cite{tariq2019gan}, \textit{FakeTalkerDetect}~\cite{jeon2019faketalkerdetect}, \textit{FaceForensics++}~\cite{rossler2018faceforensics}, and \textit{Face X-ray}~\cite{li2020face} are developed, training with a large number of real vs. forged images.
Furthermore, other researchers~\cite{marra2019gans, zhang2019detecting, yu2019attributing} have shown that the detection performance can be improved by analyzing artifacts and patterns in underlying GAN-images.

\begin{figure*}[t]
    \centering
     \includegraphics[width=0.85\linewidth]{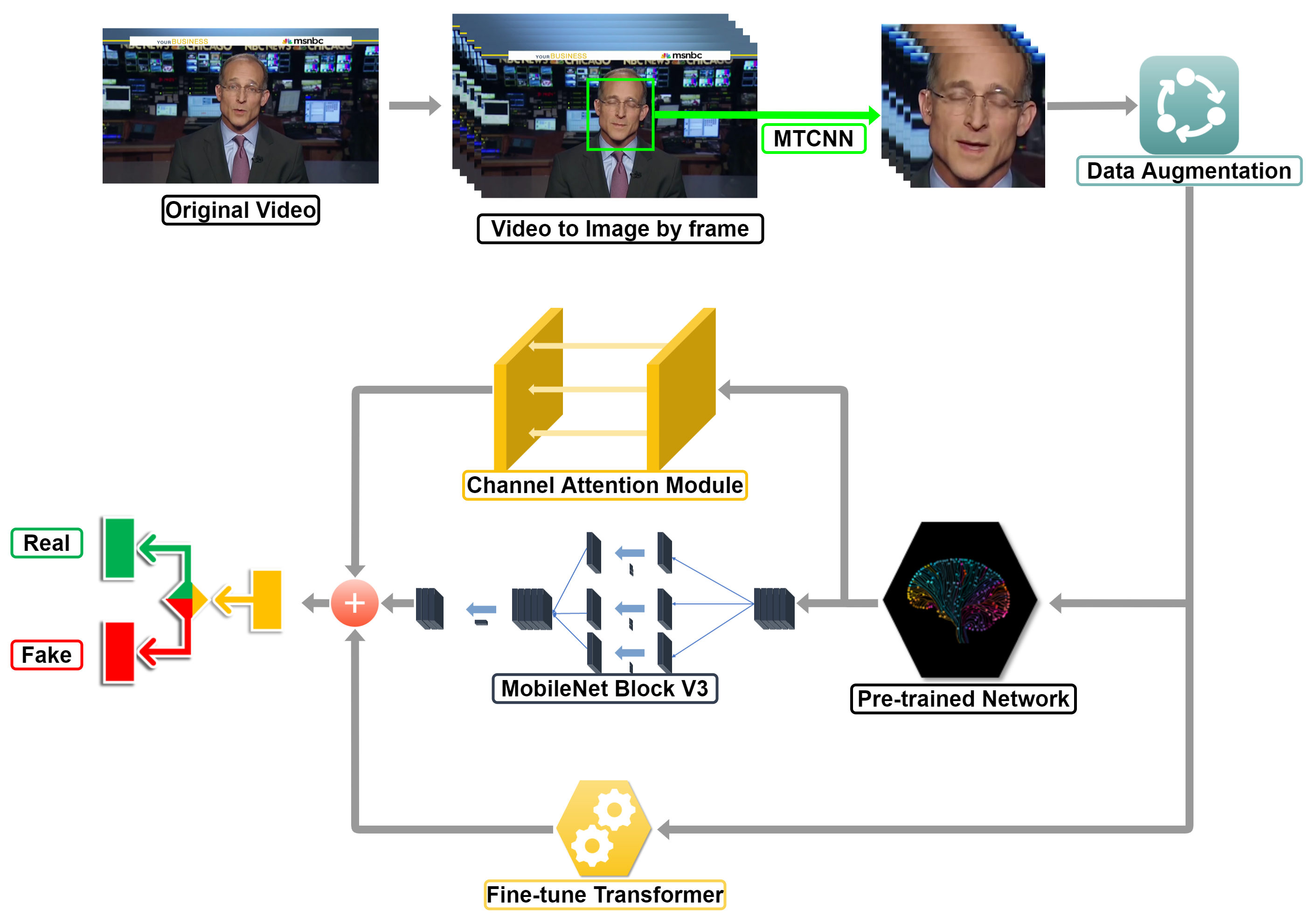}
    \caption{Overview of the DA-FDFtNet process. As our model use images as our input data, the videos are captured by frames. For a preferred classification, we use MTCNN to crop the face area of the captured frames. GAN generated images jump over the previous steps since the images already contain the face region. The images are used to pre-train the backbone model. Augmentation is applied to the image in the fine-tuning stage.}
    \label{fig:overview_fdftnet}
\end{figure*}

In this work, we proposes~\textit{Dual Attention Fake Detection Fine-tuning Network} (\textit{\SystemName}), a fine-tuning neural network-based architecture for fake face image detection. ~\SystemName~combines \textit{Fine-Tune Transformer} (FTT) and \textit{Channel attention modules} with a pre-trained Convolutional Neural Network (CNN) as a backbone, and \textit{MobileNet block V3} (MBblockV3) to distinguish the real and fake images. Figure~\ref{fig:overview_fdftnet}. presents an overview of our approach, where we implemented existing CNN architectures (pre-trained network)~\cite{simonyan2014very,szegedy2015going,he2016deep,iandola2014densenet,hu2018squeeze,howard2017mobilenets} and fine-tuning the pre-trained networks for fake image detection. In particular, \SystemName~ extends the \textit{FDFtNet}~\cite{jeon2020fdftnet}, by implementing the FTT and MBblockV3 and an additional channel attention module to the final layer. Specifically, FTT is designed to use different feature extraction from images using the self-attention. At the same time, MBblockV3 extracts the feature using different convolution and structure techniques, and the channel attention module is used to capture feature dependencies in the spatial and channel dimensions.

Moreover, various data augmentation methods have been applied to overcome the limitations of using a small dataset to fine-tune the baseline model and improve the overall performance across different types of fake images. We experiment with 4 different types of deepfakes (FaceSwap, DeepFakes, Face2Face, and NeuralTextures) as well as GAN-generated face images (StarGAN, PGGAN, StyleGAN, and StyleGAN2), and demonstrate the effectiveness of our approach across different domains. Our main contributions are summarized as follows:

\begin{itemize}
    \item We propose \SystemName, an extended neural network-based fake image detector, which achieves 97.02\% accuracy, improving the baseline model accuracy from 1\% to 47\% through our methods.
    
    \item We provide a novel fine-tuning neural network-based classifier that requires a small amount of data for fine-tuning, and can be easily integrated with popular existing CNN architectures.
    
    \item We perform an extensive evaluation 8 different deepfake as well as GAN-generated fake image datasets, and demonstrate the effectiveness of our approach over different domains.
\end{itemize}

We organized our paper as follows: In Section 2, we present an overview of fake image detection approaches and the new approach we developed for my model. In Section 3, we present the details of the dataset and baseline models, and describe the architecture of~\textit{\SystemName}. In Section 4, we explain the training details for our experiment and show our experiment results. Finally, in Section 5, we offer our conclusion. 

\section{RELATED WORK}
In this section, we provide a brief overview of prior studies 
that are directly relevant to our work.

\subsection{Traditional image forgery detection}
Previously, the frequency-domain has been explored to analyze forged fake images from real images. However, the frequency-domain based methods showed limitations in analyzing fake images with refined and smooth edges. To solve this problem, JPEG Ghost~\cite{farid2009exposing} was developed to determine the different JPEG compression quality, based on the fact that the normalized pixel distance of the reproduced image that are copied from different real images differs from the original image. Also, Error Level Analysis (ELA)~\cite{krawetz2007picture} was proposed to determine the error quality level of the images after manipulation, in which previous image forgery detection methods had shown unreasonable results.

Copy-move forgery detection~\cite{mankar2015image} was developed to classify fake images based on the pixel-based approach. First, the dyadic wavelet transform (DWT) changes the original image to the reduced dimension representation (i.e., the LL1 sub-band), where the LL1 sub-band divides into sub-images. Next, phase correlation was adopted to compute the spatial offset between the copy-move regions. Finally, Mathematical Morphological Operations (MMO) were used to remove the isolated points to improve the Copy-Move region's location. However, these digital forensic tools fail to detect the GAN-generated images, as they are entirely created from scratch, and there are no copy-forgery parts in them.

\subsection{Image forgery detection with neural networks}
ShallowNet~\cite{tariq2019gan} tackled the problem of detecting both GANs-generated human faces and human-created fake face images with neural networks using ensemble methods. Their approach provided an effective end-to-end fake face detection pipeline without resorting to any human interventions or using any metadata information. ShallowNet~\cite{tariq2019gan} outperformed previous architectures in detecting real vs. PGGAN~\cite{karras2017progressive} with a shallow layer architecture. However, their approach showed limitations when detecting other types of manipulated images, such as DeepFakes.
To help researchers better cope with different types of deepfakes, FaceForensics++~\cite{rossler2019faceforensics++} was introduced, where FaceForensics++ provides benchmark datasets and an automatic metric that takes four realistic scenarios (i.e., random encoding and dimensions). With these benchmarks, they analyzed various methods of forgery detection pipelines.
However, fine-tuning was not explored in this research. 

Yu et al.~\cite{yu2019attributing} mentioned that GAN-generated images contained a fingerprint from the generating GAN model and used an autoencoder to visualize the fingerprint and classify the real and GAN-generated fake images. Also, Nataraj et al.~\cite{nataraj2019detecting} proposed a detection method combining co-occurrence matrices and deep learning. Their method passed the co-occurrence matrices through a deep learning framework, allowing the network to learn important co-occurrence matrices essential features. Both methods~\cite{yu2019attributing, nataraj2019detecting} evaluate with only GAN generated images. However, our experiments classify both deepfake and GAN-generated images.

\subsection{Self-attention and Transformer}
Self-attention is an attention mechanism that computes the single sequence's representation by interacting with different positions from the same sequence. In fact, self-attention has been applied to various natural language processing (NLP) tasks~\cite{cheng2016long, parikh2016decomposable, paulus2017deep, lin2017structured} including machine translation. Vaswani et al.~\cite{vaswani2017attention} proposed Transformer, a model-based solely on self-attention mechanisms to draw global dependencies of inputs for machine translation.
In addition, the self-attention module was applied to the computer vision area to solve CNN models' limitation of capturing long-range and multi-level dependencies among image regions. Wang et al.~\cite{wang2018non} formalized self-attention as a non-local operation to explore the spatial-temporal dependencies' effectiveness in video and image sequences.
Parmar et al.~\cite{parmar2018image} introduced Image Transformer, applying the self-attention model into an autoregressive model for image generation.
Zhang et al.~\cite{zhang2018self} proposed SAGAN, which allowed the self-attention-driven and long-range dependency model for learning a better image generation. Hence, the generator can produce images that are blending more naturally with capturing fine-grained details through self-attention.

Fu et al.~\cite{fu2019dual} address the scene segmentation task by capturing rich contextual dependencies based on the self-attention mechanism by applying two types of attention modules on top of the dilated fully connected network.
The channel attention module selectively emphasizes inter-dependent channel maps by integrating associated features among all channel maps.
It introduces a self-attention mechanism to capture feature dependencies in the spatial and channel dimensions, respectively.

For the channel attention module, we use a similar self-attention mechanism to capture the channel dependencies between any two-channel maps and update each channel map with a weighted sum of all channel maps. \textit{\SystemName} applied the self-attention in two different ways. First, the Transformer, similar to the \textit{Multi-head Attention Module}~\cite{vaswani2017attention}, is added to Fine-Tune the network. Also, a channel attention module~\cite{fu2019dual} is used to the final layer to capture the baseline network's channel feature.
\section{Dual Attention Fake Detection Fine-tuning Network (\SystemName)}
In this section, we describe the details of our dataset and architecture design of our \SystemName. The main difference from other fake detection methods is that we utilize well-known, reusable pre-trained models and fine-tune the backbone networks with only a few data to improve the fake detection performance. Figure~\ref{fig:overview_fdftnet} shows an overview of our model, which is composed of 1) a pre-trained model, 2) Fine-Tune Transformer (FTT), 3) a MobileNet block V3 (MBblockV3), and 4) a channel attention module.

\subsection{Dataset Description}
For the face manipulation dataset, we downloaded datasets of various methods used for the experiment. FaceForensics++~\cite{rossler2019faceforensics++} provides videos generated with FaceSwap~\cite{faceswap}, DeepFakes~\cite{deepfake}, Face2Face~\cite{thies2016face2face}, and NeuralTextures~\cite{thies2019deferred}. We cropped the face region of the videos using MTCNN.  For the PGGAN~\cite{karras2017progressive}, StyleGAN~\cite{karras2019style}, and StyleGAN2~\cite{karras2020analyzing} images, we used the official dataset provided. Since StarGAN~\cite{choi2018stargan} does not provide the generated dataset, we generated the images following the official source code presented.

\begin{figure}[h!]
\begin{center}
   \includegraphics[width=1.0\linewidth]{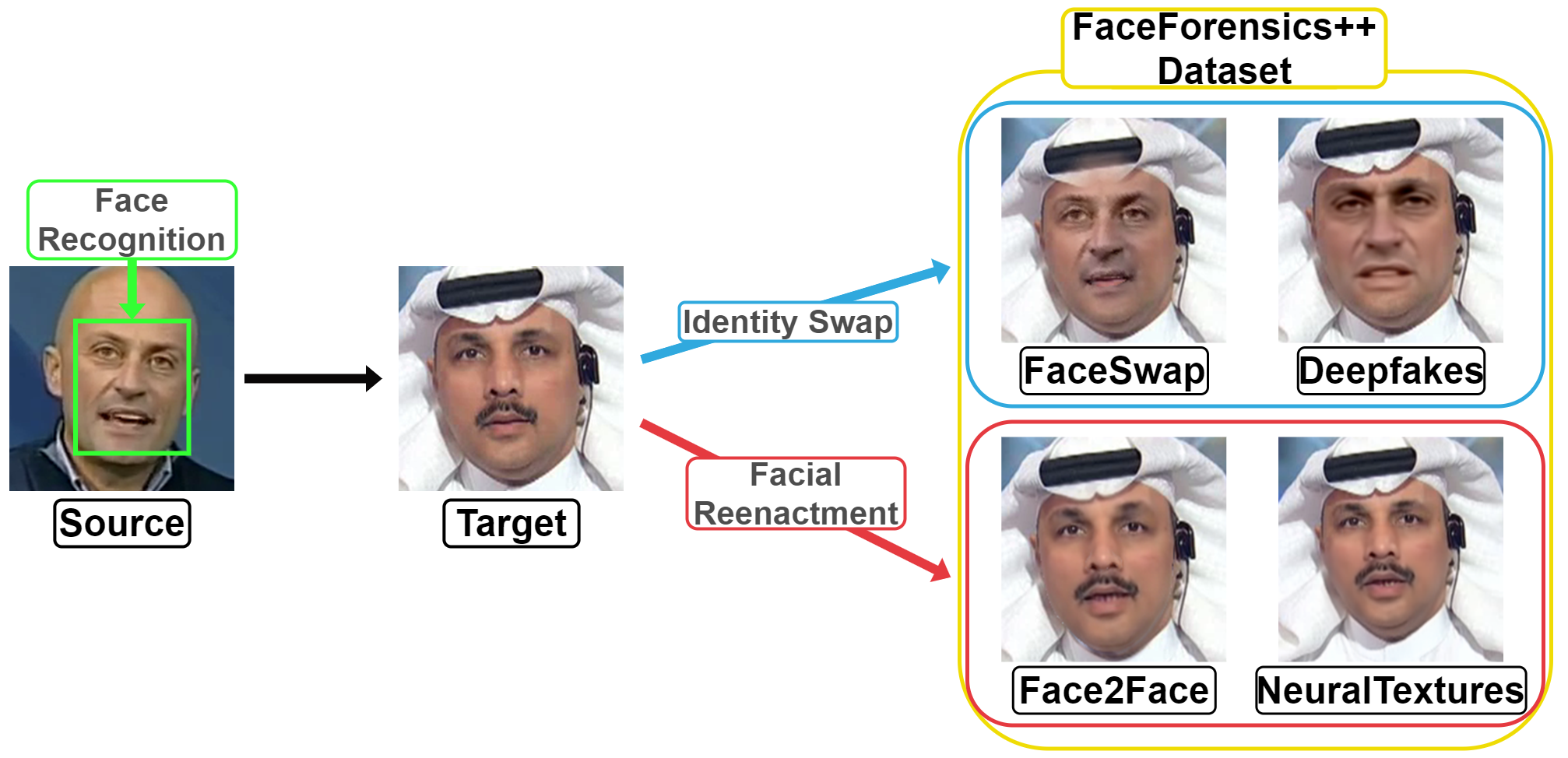}
\end{center}
   \caption{FaceForensics++ contains manipulated videos for the research of deep learning-based approaches. It is created with for methods, FaceSwap, DeepFakes, Face2Face, and NeuralTextures.}
\label{fig:faceforensics++_dataset}
\end{figure}

\noindent\textbf{FaceSwap.}
In our experiment, we used the dataset provided by FaceForensics++~\cite{rossler2019faceforensics++}, which implements a GitHub repository\footnote{\url{https://github.com/MarekKowalski/FaceSwap/}}. FaceSwap~\cite{faceswap} changes the face of a target person to the source image face. To apply the FaceSwap model, the face region and landmarks of the input image fit the 3D model. The 3D model is organized with vertices of a neutral face, blendshapes, a set of triplets to form the mesh of the face, and indices that correspond between the landmarks from the localizer and the vertices of the 3D face shape. The texture coordinates use the vertices from the 3D model landmarks. After the input image step, the model locates the target image's face region and facial landmarks captured from the target video. The 3D model of the input image is fitted to the located landmarks and renders the swapped face. The image is blended smoothly to produce the final output image.

\noindent\textbf{DeepFakes.}
Recently Deepfakes~\cite{deepfake} has become a word that represents most of the face manipulated images based on deep learning. Various kinds of Deepfakes are available from different providers. For our experiment, we downloaded the dataset provided by FaceForensics++~\cite{rossler2019faceforensics++}, denoted as DeepFakes. FaceForenscis++ implements the faceswap GitHub~\cite{faceswap} to generate the DeepFakes dataset. The face of the observed source video replaces the face of the target video. Two autoencoders share an encoder trained to reconstruct the face of the training images from the source and target. A face detector crops and rearranges the face of the images. The trained encoder and the source face decoder are applied to the target face to create a fake image.

\noindent\textbf{Face2Face.}
Face2Face~\cite{thies2016face2face} is a facial reenactment scheme that transfers a source video, using the facial expressions from the target video and re-render the manipulated output to a realistic fake video. At run time, both source and target video's facial expressions are tracked using a dense photometric consistency measure. The best matching mouth interior is retrieved from the target sequence to produce more accurate fit. The corresponding video re-render the synthesized target face producing a final composite fake video.

\noindent\textbf{NeuralTextures.}
NeuralTextures proposed by Thies et al.~\cite{thies2019deferred} include the learned feature maps which are trained as part of the scene capture process.
The difference with Neural Textures compared to traditional textures is that the stored high-dimensional feature maps contain more information.
The FaceForensics++~\cite{rossler2019faceforensics++} implements the idea and uses the original video data to learn a neural texture of the target person, including a rendering network.
The difference is that the FaceForensics++ applies a patch-based GAN-loss from Pix2Pix~\cite{isola2017image}.
The tracking module of Face2Face~\cite{thies2016face2face} is used for the geometry of train and test times of the NeuralTextures dataset~\cite{rossler2019faceforensics++}.

\begin{figure}[h!]
\begin{center}
   \includegraphics[width=1.0\linewidth]{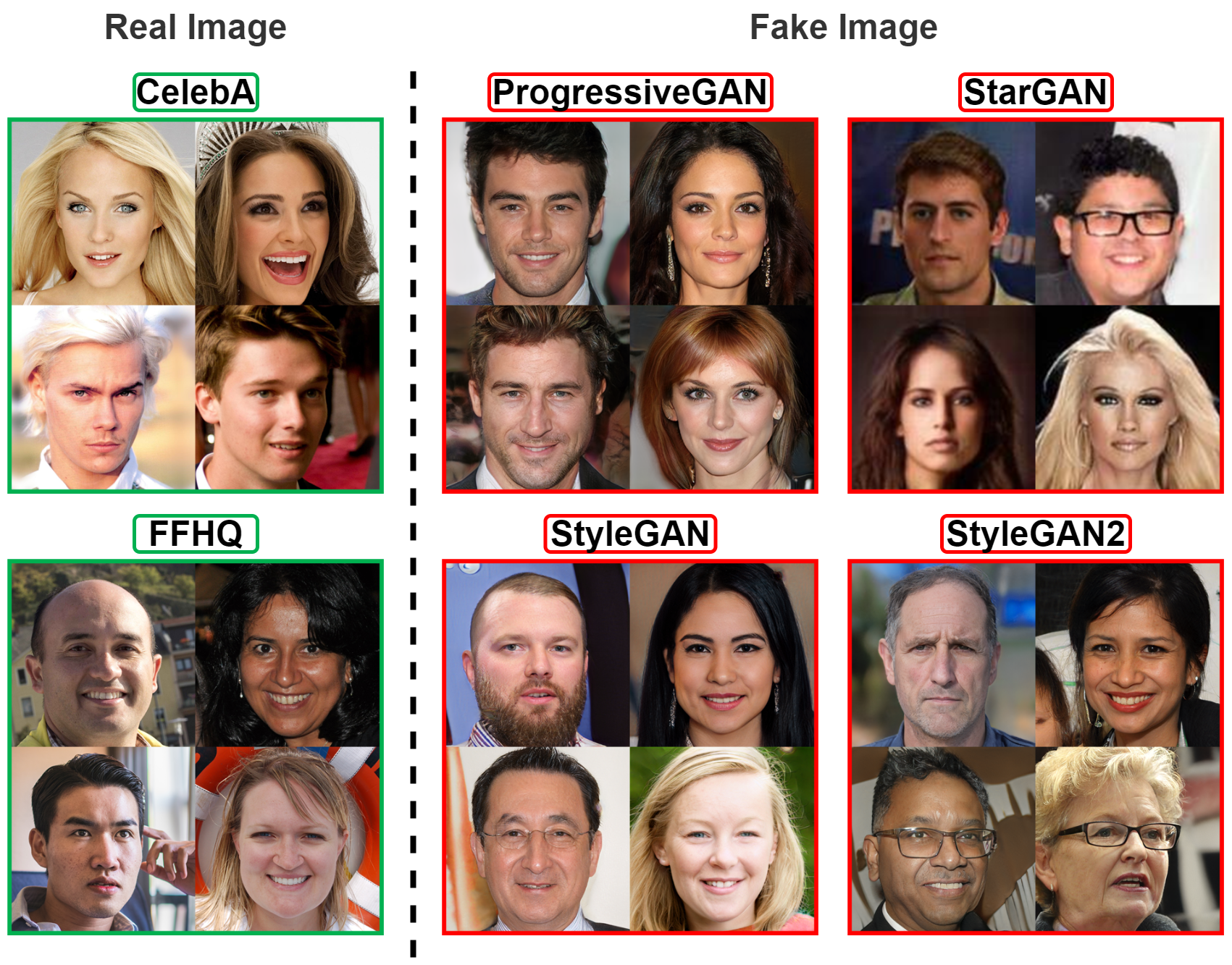}
\end{center}
   \caption{GAN generated images used for our experiment. Progressive growing GAN and StarGAN both used the CelebA dataset for training. StyleGAN and StyleGAN2 are generated from the FFHQ dataset.}
\label{fig:GAN_dataset}
\end{figure}

\noindent\textbf{PGGAN.}
The key idea of Progressive growing GAN~\cite{karras2017progressive} is to grow the generator and discriminator progressively.
While the model starts from the low-resolution images, new layers are added to the model as training progress, which speeds the training and stabilizes the model to generate high-resolution images.
The CelebA-HQ dataset, processed from the CelabA dataset~\cite{liu2015deep}, is used for the input datasets for PGGAN~\cite{karras2017progressive} to provide 100,000 GAN-generated fake celebrity images.

\noindent\textbf{StarGAN.}
StarGAN~\cite{choi2018stargan} performs image-to-image translations by training data of multiple domains and learns the mappings between all available domains using only a single model.
Instead of learning a fixed translation, the generator inputs both image and domain information and learns to translate the image into the corresponding domain flexibly.
This allows StarGAN~\cite{choi2018stargan} to simultaneously train the multiple datasets with different domains within a single network, leading to a quality of translated images, especially facial attribute transfer and facial expression synthesis tasks.

\noindent\textbf{StyleGAN.}
StyleGAN~\cite{karras2019style} architecture leads to an automatically learned, unsupervised separation of high-level attributes and stochastic variation in the generated images, enabling intuitive and scale-specific control of the synthesis process.
For StyleGAN-images, we used the official implementation dataset\footnote{\url{https://github.com/NVlabs/stylegan}} provided by the author, consisting of 100,000 GAN-generated celebrity images at a 1024$\times$1024 resolution generated from the FFHQ dataset~\cite{karras2019style}.
For our experiment, we resized the image to a 256$\times$256 resolution.

\noindent\textbf{StyleGAN2.}
StyleGAN2~\cite{karras2020analyzing} redesigns the generator normalization, revisits the progressive growth, and regularizes the generator to encourage good conditioning when mapping latent vectors to images.
For StyleGAN2-images, we used the official implementation dataset\footnote{\url{https://github.com/NVlabs/stylegan2}} provided by the author, under the same condition as in StyleGAN~\cite{karras2019style}.

\subsection{Description of Pre-trained Backbone CNN networks}
We used the following CNN networks as our backbone networks, as shown in Fig.~\ref{fig:overview_fdftnet}, and our baselines (backbone networks): SqueezeNet~\cite{iandola2016squeezenet}, ShallowNetV3~\cite{tariq2019gan}, and Xception~\cite{chollet2017xception}.
Each network is pre-trained from each dataset (i.e., FaceSwap, DeepFakes~\cite{deepfake}, Face2Face~\cite{thies2016face2face}, NeuralTextures~\cite{thies2019deferred}, PGGAN~\cite{karras2017progressive}, StarGAN~\cite{choi2018stargan}, StyleGAN~\cite{karras2019style}, and StyleGAN2~\cite{karras2020analyzing}).

\noindent \textbf{SqueezeNet.}
Smaller CNN architectures require less communication during distributed training, less bandwidth to export a new model and more feasible deployments to hardware.
To apply this, SqueezeNet~\cite{iandola2016squeezenet} decreased the number of parameters in the CNN architecture and maintained the accuracy of AlexNet~\cite{krizhevsky2012imagenet}
We chose SqueezeNet~\cite{iandola2016squeezenet} as our baseline because Squeezenet~\cite{iandola2016squeezenet} is not suitable for fake detection, thus will show a low accuracy.
However, applying our fine-tune method yields an improvement inaccuracy.

\noindent \textbf{ShallowNetV3.}
The purpose of ShallowNet~\cite{tariq2019gan} is to classify both GANs and human-created fake face images without restoring the metadata information.
ShallowNetV3~\cite{tariq2019gan} showed meaningful results on the CelebA~\cite{liu2015deep} and Progressive GAN~\cite{karras2017progressive} classification task. For the real-world scenario, we tested the ShallowNet~\cite{tariq2019gan} not only on GAN generated images but also on various face manipulation methods.

\noindent \textbf{Xception.}  
For the detection accuracy, FaceForensics++~\cite{rossler2019faceforensics++} implemented Xception~\cite{chollet2017xception} as the baseline model. Xception~\cite{chollet2017xception} replaces the Inception modules in the Inception model~\cite{szegedy2016rethinking} with depthwise separable convolutions, making the Xception~\cite{chollet2017xception} model free from FC layers. The number of parameters is maintained while the performance increases.

\subsection{Fine-Tune Transformer (FTT)}
\noindent \textbf{Self-Attention Module.}
Our Fine-Tune Transformer (FTT) is organized with multiple self-attention modules as presented in Figure~\ref{fig:self_attention}, where the self-attention module's main goal is to determine where to focus on the features to discriminate between real and fake input images. Each self-attention module has a 1$\times$1 convolution filtering process to obtain the feature spaces $f$($x$), $g$($x$), and $h$($x$) of the input image $x$. Equation~\ref{equation:transformer_conv}. shows the convolution filtering step, where $W_f, W_g$, and $W_h$ are the respective filter weights of each space. To obtain the attention map $\alpha$ in Figure~\ref{fig:self_attention}, the result of the Softmax operation of the $i^{th}$ feature space and $j^{th}$ feature space is calculated using the dot-product attention in Equation~\ref{equation:transformer_softmax} as follows:

\begin{figure}[h!]
\begin{center}
  \includegraphics[width=0.65\linewidth]{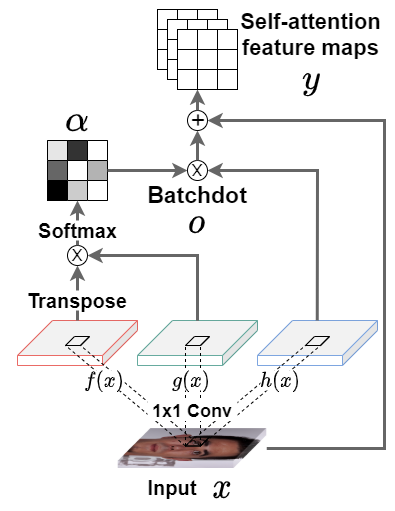}
\end{center}
  \caption{Self-attention module in the Fine-Tune Transformer. The input $x$ goes through a $1\times1$ convolution filter into $f$, $g$, and $h$. The attention map $\alpha$ is the softmax result of $f$ and $g$. The batchdot $o$ multiplies $h$ and the attention map $\alpha$. The input image $x$ is added to the result $o$. The final output $y$ is the self-attention feature maps.}
\label{fig:self_attention}
\end{figure}

\begin{align}
    f\left(x\right) &\text{ = } W_f  x, \indent g\left(x\right) \text{ = } W_g  x, \indent h\left(x\right) \text{ = } W_h x
    \label{equation:transformer_conv}
    \\
    \alpha_{j,i} &\text{ = } Softmax\left(f\left(x_i\right)^T \text{ , }   g\left(x_j\right)\right).
    \label{equation:transformer_softmax}
\end{align}

\noindent After obtaining the attention map $\alpha$, the $Batchdot$ operation is applied to multiply the attention map $\alpha_{j,i}$ with $h$($x$), and the output $o_j$ is the attention of the input image, as shown in Equation~\ref{equation:batch_dot}. Also, we multiply $\gamma$ with attention $o_j$ and then add $\gamma$ $o_j$ to the input $x_i$, as shown in Equation~\ref{equation:gamma}, to obtain the final self-attention feature map $y_i$. In particular, $\gamma$ is a learnable parameter initialized as 0 at the early stage of learning. This is suitable since the softmax function equally provides attention to all the feature spaces at the early learning stage.

\begin{align}
    o_j & \text{ = } Batchdot\left(\alpha_{j,i} \text{ , } h\left(x\right)\right) \label{equation:batch_dot}
    \\
    y_i & \text{ = } \gamma o_j \text{ + } x_i 
    \label{equation:gamma}
\end{align}

\begin{figure}[h!]
\begin{center}
  \includegraphics[width=0.8\linewidth]{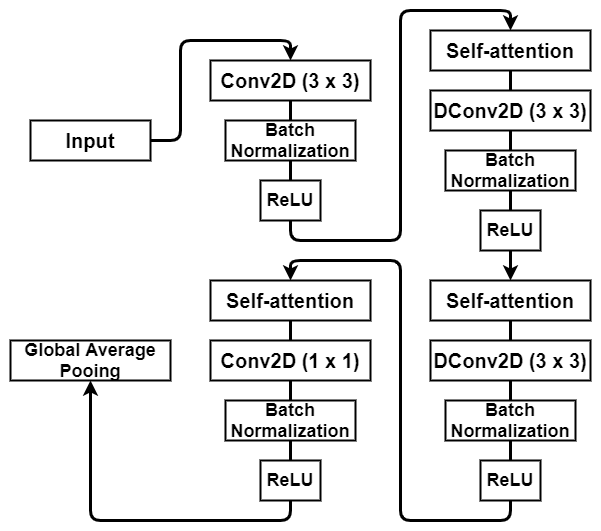}
\end{center}
  \caption{Specification for Fine-Tune Transformer (FTT). Conv and DConv each denote convolution, depth-wise separable convolution, respectively. We repeat FTT three times ($M$ = $3$) to maximize the performance.}
\label{fig:finetune}
\end{figure}

\noindent \textbf{Transformer.}
As shown in Figure~\ref{fig:finetune}, we use the Transformer architecture~\cite{vaswani2017attention} to overcome the problem of CNN having limited receptive fields. The receptive fields cause problems in achieving long-term dependencies by numerous Convolution filters with a small size. In particular, we apply the self-attention module three times, as shown in Figure~\ref{fig:finetune}, where first layer is a 3$\times$3 separable convolution followed by Batch Normalization (BN)~\cite{ioffe2015batch} and ReLU. After that, self-attention is performed three times ($M$ = $3$), followed by SeparableConv 3$\times$3, BN, and ReLU. The dimension of the output feature map from the self-attention module is 32, 64, and 128, respectively.

\subsection{Channel Attention}

\begin{figure}[h!]
\begin{center}
  \includegraphics[width=1\linewidth]{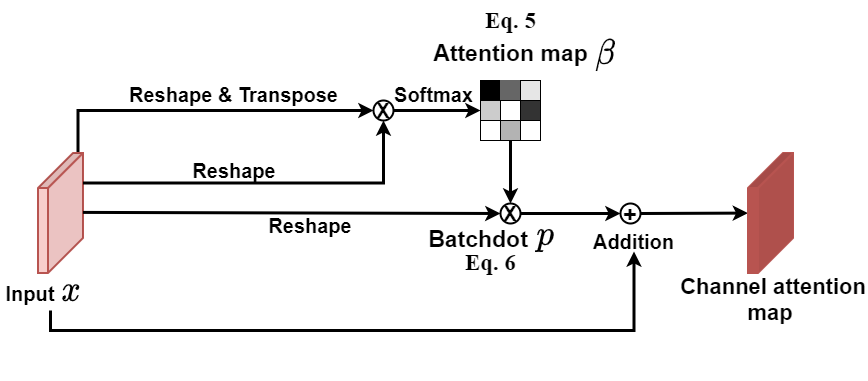}
\end{center}
  \caption{Details of the channel-attention module. The input $x$ (the image or the output from the previous layer) is reshaped. The attention map $\beta$ is the softmax output from the softmax result from the reshaped input. The Batchdot $o$ multiplies the reshaped input and the attention map $\beta$. The input image $x$ is added to $o$. The final output $y$ is the channel attention feature map.}
\label{fig:channel_attention}
\end{figure}

\noindent Figure~\ref{fig:channel_attention} illustrates the structure of the channel attention module. Unlike FTT self-attention module in Figure ~\ref{fig:self_attention}, the original feature $x$ directly calculates the channel attention map, where the input $x$ is reshaped to a $C$$\times$$N$ shape where $N$ is the number of pixels H$\times$W. The reshaped input $x$ is matrix multiplied with the transpose of itself and applies a softmax layer to integrate the attention map $\beta$ in Equation~\ref{equation:channel_softmax}. Similar to Figure~\ref{fig:self_attention}, the Batchdot operation is applied to attention map $\beta$ and the reshaped input $x$ in Equation~\ref{equation:channel_batch_dot}. To obtain the final result, we add the result of the Batchdot operation and the input $x$ as follows:

\begin{align}
    \beta_{j,i} &\text{ = } Softmax((x_i)^T \text{ , } x_j)
    \label{equation:channel_softmax}
    \\
    p_j & \text{ = } Batchdot\left(\beta_{j,i}\text{ , } x\right) \label{equation:channel_batch_dot}
\end{align}

\subsection{MobileNet block V3}
Our model used MBblockV3~\cite{howard2019searching} to be repeated before the classification layer to extract the feature space over the pre-trained feature space. MobileNetV1~\cite{howard2017mobilenets} provided the depth-wise convolution to reduce the model size and the number of parameters. MobileNetV2~\cite{sandler2018mobilenetv2} proposed an additional Inverted Residual Block expansion layer in the block to help reduce memory usage and improve performance. Finally, for MobileNetV3, squeeze and excitation layers were added in the initial building block taken from MobileNetV2. The addition of the squeeze and excitation~\cite{hu2018squeeze} modules in MobileNetV3 improved the concentrating on the largest representation of the extended features.

\begin{figure}[h!]
\begin{center}
  \includegraphics[width=0.9\linewidth]{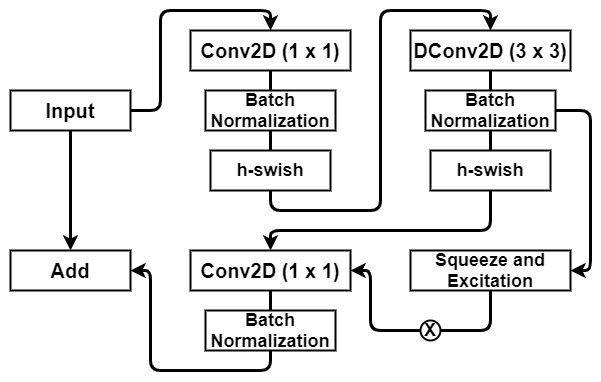}
\end{center}
  \caption{Specification for MBblockV3. Conv, BN, DConv, and GAP denote convolution, batch normalization, depth-wise separable convolution, and global average pooling. If the stride of 3x3 DConv is 2, the addition operation is skipped, and $W$ and $H$ are divided by 2. Bold operations represent the Squeeze-and-Excitation block.}
\label{fig:mobilenetblockv3}
\end{figure}

For repetition $N$, we use $N=4$, which returned the best fine-tuning results. We use the modified h-swish and the ReLU6 as activation functions on the top layers to reduce the distortion in the data distribution, and extract different signals from the ReLU layer layer represented in Equation~\ref{equation:hswish}.

In the squeeze stage, the global information on the image resolution is embedded to the extracted the Squeeze-and-Excitation~\cite{hu2018squeeze} blocks. Then, the gathered information is used to capture channel dependencies and re-calibrated with the gated computation (element-wise multiplication).

\begin{equation}
\begin{split}
    \label{equation:hswish}
        \text{ReLU6[x]} \text{ = } min\left(max\left(0, x\right), 6 \right),\\
        \text{h-swish[x]} \text{ = } x\frac{\text{ReLU6} \left (x \text{ + } 3 \right)}{6}
\end{split}
\end{equation}
\section{Experimental Results}
\subsection{Training details}
All datasets have train, validation, test, fine-tune, and fine-tune sets. The size of each dataset is shown in Table~\ref{table:volume}. To verify our model's efficiency, we only used 2,000 images for the fine-tuning, which is small compared to the 60,000 images for baseline training. \SystemName~is trained with SGD and Adam optimizer~\cite{kingma2014adam} depending on the dataset for 200 epochs. The mini-batch size 64 is used; however, different mini-batch sizes can be applied flexibly depending on the memory of the environment, and early stopping is applied when the validation loss ceases to decrease for 20 epochs. All input images are resized to 64$\times$64 resolution to reenact the most challenging scenarios in detecting fake images.

\begin{table*}[h!]
\renewcommand{\arraystretch}{1.0}
\caption{The respective size of the train, validation, test, and fine-tune sets. We only use each 1,000 real and fake images, respectively, for fine-tuning.}
\label{table:volume}
\begin{center}
     \resizebox{\textwidth}{!}{
    \small
    \begin{tabular}{l||c|c|c|c|c|c|c|c}
        \hline
        \ Dataset & FaceSwap & DeepFakes & Face2Face & NeuralTextures & PGGAN & StarGAN & StyleGAN & StyleGAN2 \\
        \hline
        \ Train & 60,000 & 60,000 & 60,000 & 60,000 & 60,000 & 60,000 & 60,000 & 60,000 \\
        \hline
        \ Validation & 18,000 & 18,000 & 18,000 & 18,000 & 18,000 & 18,000 & 18,000 & 18,000\\
        \hline
        \ Test & 20,000 & 20,000 & 20,000 & 20,000 & 20,000 & 20,000 & 20,000 & 20,000\\
        \hline
         \ \textbf{Fine-tune} & \textbf{2,000} & \textbf{2,000} & \textbf{2,000} & \textbf{2,000} & \textbf{2,000} & \textbf{2,000} & \textbf{2,000} & \textbf{2,000}\\
        \hline
    \end{tabular}}
\end{center}
\end{table*}

\subsection{Data Augmentation}

In order to improve the performance, we apply different data augmentation approaches in this section.

\begin{figure}[h!]
\centering
    \includegraphics[width=0.9\linewidth]{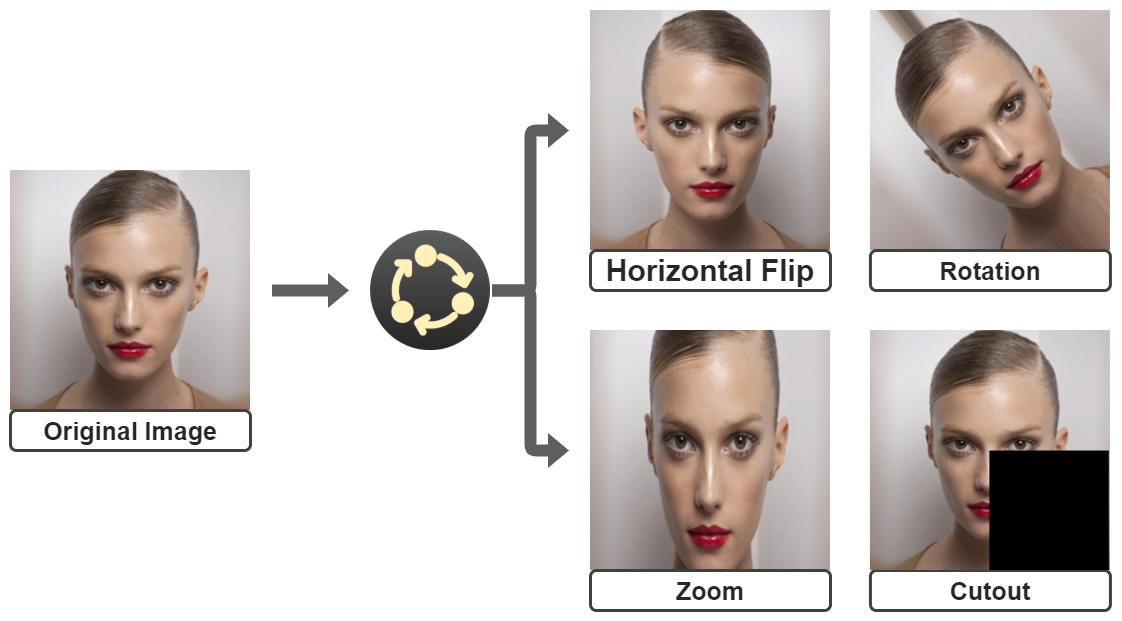}
\caption{Examples of the data augmentations used in our fine-tune training. Each ratio was randomly chosen for augmentation.}
\label{fig:cutout}
\end{figure}

\noindent\textbf{Cutout.}
We applied the Cutout by DeVries et al.~\cite{devries2017improved}, which covers the input image's random location with a squared zero mask.
In the original paper, Devries et al.~\cite{devries2017improved} used random zero masks of 16 pixels for CIFAR-10 (32$\times$32 pixels images), 5 random iteration parameters $\alpha$ for cutting, and 16 random size multipliers $\beta$ with random center cropping to cut the masks.
However, implementing the original setting caused underfitting in our experiment.
We obtained higher performance when using 4$\times$4 pixel masks, 3 iterations, 5-size multipliers, and random translations instead of random center cropping to cut the masks for 64$\times$64 images ($\alpha$ = 3 and $\beta$ = 5).
Examples of the Cutout method is shown in Figure~\ref{fig:cutout}.

\begin{table*}[h!]
\begin{center}
\caption{Overall performance evaluation results of the FaceForensics++ dataset. The evaluation metrics used are ACC (\%) and AUROC (\%). The underlined results are improved performance compared to the baseline, and the best detection results among all are highlighted in bold.}
\label{table:deepfake_performance}
    \resizebox{\textwidth}{!}{
    \begin{tabular}{l||c||cc||cc||cc||cc}
    \hline
    \multirow{2}{*}{Model} &
    \multirow{2}{*}\ Dataset & \multicolumn{2}{c}{FaceSwap} & \multicolumn{2}{c}{DeepFakes} & \multicolumn{2}{c}{Face2Face} & \multicolumn{2}{c}{NeuralTextures}\\\cline{2-10} & Backbone & ACC (\%) & AUROC & ACC (\%) & AUROC  & ACC (\%) & AUROC  & ACC (\%) & AUROC\\
    \hline
    SqueezeNet & baseline & 50.00 & 50.00 & 50.00 & 50.00 & 50.00 & 50.00 & \underline{98.80} & 99.52\\
    \SystemName \space (Ours) & SqueezeNet & \underline{60.76} & \underline{65.89} & \underline{78.83} & \underline{90.01} & \underline{73.94} & \underline{83.43} & 97.20 & \textbf{99.77}\\
    \hline
    ShallowNetV3$\dagger$& baseline & 82.81 & 84.96 & 50.00 & 50.00 & 93.68 & 97.49 & 98.70 & 99.70\\
    \SystemName \space (Ours) & ShallowNetV3 & \underline{83.93} & \underline{85.53} & \underline{73.06} & \underline{88.01} & \underline{91.33} & \underline{97.26} & \textbf{98.75} & \underline{99.57}\\
    \hline
    Xception & baseline & 88.60 & 95.35 & 91.24 & 97.04 & 87.80 & 94.72 & \underline{99.09} & \underline{99.96}\\
    \SystemName \space (Ours) & Xception & 
    \textbf{90.51} & \textbf{95.83} &
    \textbf{97.69} & \textbf{99.49} &
    \textbf{92.60} & \textbf{97.30} &
    97.45 & 99.63\\
    \hline
    \end{tabular}}
\end{center}
\end{table*}

\begin{table*}[h!]
\begin{center}
\caption{Overall performance evaluation results of the GAN dataset. The evaluation metrics used are ACC (\%) and AUROC (\%). The underlined results are improved performance compared to the baseline, and the best detection results among all are highlighted in bold.}
\label{table:gan_performance}
    \resizebox{\textwidth}{!}{
        \begin{tabular}{l||c||cc||cc||cc||cc}
    \hline
    \multirow{2}{*}{Model} &
    \multirow{2}{*}\ Dataset & \multicolumn{2}{c}{StarGAN} & \multicolumn{2}{c}{PGGAN} & \multicolumn{2}{c}{StyleGAN} & \multicolumn{2}{c}{StyleGAN2}\\\cline{2-10} & Backbone & ACC (\%) & AUROC & ACC (\%) & AUROC  & ACC (\%) & AUROC  & ACC (\%) & AUROC\\
    \hline
    SqueezeNet & baseline & 50.00 & 50.00 & 50.00 & 50.00 & 50.00 & 50.00 & 50.00 & 50.00\\
    \SystemName \space (Ours) & SqueezeNet & \underline{65.63} & \underline{71.19} & \underline{91.54} & \underline{97.18} & \underline{58.47} & \underline{61.67} & \underline{57.65} & \underline{62.71}\\
    \hline
    ShallowNetV3$\dagger$& baseline & 85.73 & 92.90 & 72.79 & 73.55 & 77.75 & 77.24 & 69.30 & 72.64\\
    \SystemName \space (Ours) & ShallowNetV3 & \underline{88.03} & \underline{94.53} & \underline{75.08} & \underline{75.67} & \underline{84.05} & \textbf{91.02} & \textbf{85.35} & \underline{82.28}\\
    \hline
    Xception & baseline & 87.12 & 94.96 & 93.53 & 82.90 & 81.77 & 83.51 & 80.82 & 82.61\\
    \SystemName \space (Ours) & Xception & \textbf{90.29} & \textbf{95.98} &
    \textbf{97.83} & \textbf{98.25} &
    \textbf{81.43} & \underline{83.11} &
    \underline{83.64} & \textbf{84.38}\\
    \hline
    \end{tabular}}
\end{center}
\end{table*}

\subsection{Performance evaluation}
We present our overall performance results in Table~\ref{table:deepfake_performance}, and Table~\ref{table:gan_performance}, respectively. We use the accuracy (ACC) and AUROC as evaluation metrics. We experimented with all three baseline models: SqueezeNet~\cite{iandola2016squeezenet}, ShallowNetV3~\cite{tariq2019gan}, and Xception~\cite{chollet2017xception}, on each dataset with similar training strategies. The experimental results show that our \SystemName~improves the detection performance in both ACC and AUROC compared to all the baselines. Our model shows high performance using 1,000 images for real and fake in terms of training data size, respectively.

\noindent\textbf{FaceForensics++ dataset.}
The weight parameters of the pre-trained models are frozen to achieve the best performance. We applied $\alpha = 3$ iterations and $\beta = 5$ multipliers for the Cutout parameters for the data augmentation. For the FTT, we used $M = 3$ for the self-attention iteration and $N = 4$ for the stack of MBblockV3. Results demonstrate that most of the models achieved performance improvement. Table~\ref{table:deepfake_performance}. shows that Xception reached the highest accuracy and AUROC score in all the baseline models. At the same time, SqueezeNet showed the lowest score in all the datasets, with a 50.00\% accuracy AUROC score. Applying our approach, the baseline model accuracy and AUROC score increased from a minimum of 1\% to a maximum of 40\%, depending on the dataset and baseline model. The DeepFakes dataset detection with SqueezeNet showed the highest increase rate, 50.00\% ACC to 78.83\% ACC, and 50.00\% AUROC to 90.01\% AUROC. Experiments on the NeuralTexture dataset showed a slight increase due to the excessively high baseline scores. \SystemName~accomplished the highest ACC and AUROC scores for each dataset with Xception as the baseline model.

\noindent\textbf{GAN generated dataset.}
The training strategies for the GAN-generated images dataset are the same as those of the FaceForensics++ dataset. The same data augmentation is applied to the fine-tuning data with the Cutout parameters $M$, $N$, $\alpha$, and $\beta$ set to $3$, $4$, $3$, and $10$, respectively. Once more, SqueezeNet achieved the lowest score, 50.00\% ACC, and 50.00\% AUROC on all the datasets. However, the interesting point is that applying our \SystemName~with the SqueezeNet baseline again produced a high increase in both ACC and AUROC scores, approximately around 40\% with the PGGAN dataset. In general, \SystemName~increases the baseline performance from a minimum of 2\% to a maximum of 40\%. For the Xception model, our \SystemName~clearly increased the accuracy of the baseline Xception model compared to other baseline models.

\subsection{Ablation Study}
We estimate the efficiency of the channel attention module through an ablation study. For the results of Table~\ref{table:ablation}, we chose SqueezeNet model as our baseline model. We used DeepFakes and StarGAN,  each from the FaceForensics++ and GAN image dataset, to compare the results.
We compare our \SystemName~with the original FDFtNet~\cite{jeon2019faketalkerdetect}, which does not include the channel attention module to capture the channel features. We used the same hyperparameter settings from the previous experiment for both FDFtNet and~\SystemName.
Applying the channel attention module, we increased the performance by about an average of 2.3\% compared to the previous FDFtNet.

\begin{table}[h!]
\begin{center}
\renewcommand{\arraystretch}{1.0}
\caption{Performance of the Ablation study.~\SystemName~contains an additional channel attention module compared to the FDFtNet. The bold results mark out the best detection results.}
\resizebox{\columnwidth}{!}{%
\label{table:ablation}
    \scalebox{1.0}{
    \Huge
    \begin{tabular}{l||c||cc||cc}
    \hline
    \multirow{2}{*}{Model} &
    \multirow{2}{*}\ Dataset & \multicolumn{2}{c}{DeepFakes} & \multicolumn{2}{c}{StarGAN} \\\cline{2-6} & Backbone & ACC (\%) & AUROC & ACC (\%) & AUROC\\
    \hline
    FDFtNet & SqueezeNet & 75.12 & 87.06 & 63.44 & 69.66\\
    \SystemName \space (Ours) & SqueezeNet & \textbf{78.83} & \textbf{90.01} & \textbf{65.53} & \textbf{71.19}\\
    \hline
    \end{tabular}}}
\end{center}
\end{table}

\section{Conclusion}

We propose~\SystemName, to detect fake images by fine-tuning the pre-trained CNN architectures. Our experimental results show that our approach outperforms over other baselines by integrating FTT, MBblockV3, and a channel attention module with only using additional 1,000 new real and 1,000 new fake images. Our~\SystemName shows good results in both FaceForensics++ and GAN generated dataset. Therefore,~\SystemName~can be an effective choice for detecting various fake images in a real-world scenario, where available new fake images are extremely small or difficult to collect.



\section{Acknowledgements}
This work was supported by Institute of Information \& communications Technology Planning \& Evaluation(IITP) grant funded by the Korea government (MSIT) (No.2019-0-00421, Artificial Intelligence Graduate School Program (Sungkyunkwan University)).


 \bibliographystyle{elsarticle-num} 
 \bibliography{reference.bib}





\end{document}